\pdfoutput=1

\documentclass[11pt]{article}

\usepackage{acl}

\usepackage{times}
\usepackage{latexsym}
\usepackage{multirow}
\usepackage{tabularx}
\usepackage{booktabs}

\usepackage[T1]{fontenc}

\usepackage[utf8]{inputenc}

\usepackage{microtype}

\usepackage{graphicx}
\usepackage{amsmath}

\title{EmpHi: Generating Empathetic Responses with Human-like Intents}

\newcommand*\samethanks[1][\value{footnote}]{\footnotemark[#1]}
\author{Mao Yan Chen\thanks{\quad Equal contribution.} 
       \and Siheng Li\samethanks
       \and YujiuYang\thanks{\quad Corresponding author(\href{mailto: yang.yujiu@sz.tsinghua.edu.cn}{yang.yujiu@sz.tsinghua.edu.cn}).} \\
       Shenzhen International Graduate School, Tsinghua University \\
       \{chenmaoy19, lisiheng21\}@mails.tsinghua.edu.cn \\
       yang.yujiu@sz.tsinghua.edu.cn
       }

\begin{document}
\maketitle
\begin{abstract}
In empathetic conversations, humans express their empathy to others with empathetic intents. However, most existing empathetic conversational methods suffer from a lack of empathetic intents, which leads to monotonous empathy. 
To address the bias of the empathetic intents distribution between empathetic dialogue models and humans, we propose a novel model to generate \textbf{emp}athetic responses with \textbf{h}uman-consistent empathetic \textbf{i}ntents, \textbf{EmpHi} for short. 
Precisely, EmpHi learns the distribution of potential empathetic intents with a discrete latent variable, then combines both implicit and explicit intent representation to generate responses with various empathetic intents. 
Experiments show that EmpHi outperforms state-of-the-art models in terms of empathy, relevance, and diversity on both automatic and human evaluation. 
Moreover, the case studies demonstrate the high interpretability and outstanding performance of our model. Our code are avaliable at \href{https://github.com/mattc95/EmpHi}{https://github.com/mattc95/EmpHi}. 
\end{abstract}

\section{Introduction}
Empathy is a basic yet essential human ability in our daily life. It is a capacity to show one's caring and understanding to others. 
Many types of research have been conducted on empathetic expression to enhance the empathy ability of Artificial Intelligence, e.g., computational approach for empathy measurement \cite{sharma-etal-2020-computational}, empathetic expression understanding in newswire \cite{buechel-etal-2018-modeling}, online mental health support \cite{sharma2021facilitating}, etc. 
In this work, we focus on the task of generating empathetic responses \cite{rashkin-etal-2019-towards, lin-etal-2019-moel, majumder-etal-2020-mime} in open-domain conversation.

\begin{figure}[tp]
    \includegraphics[width={\linewidth}]{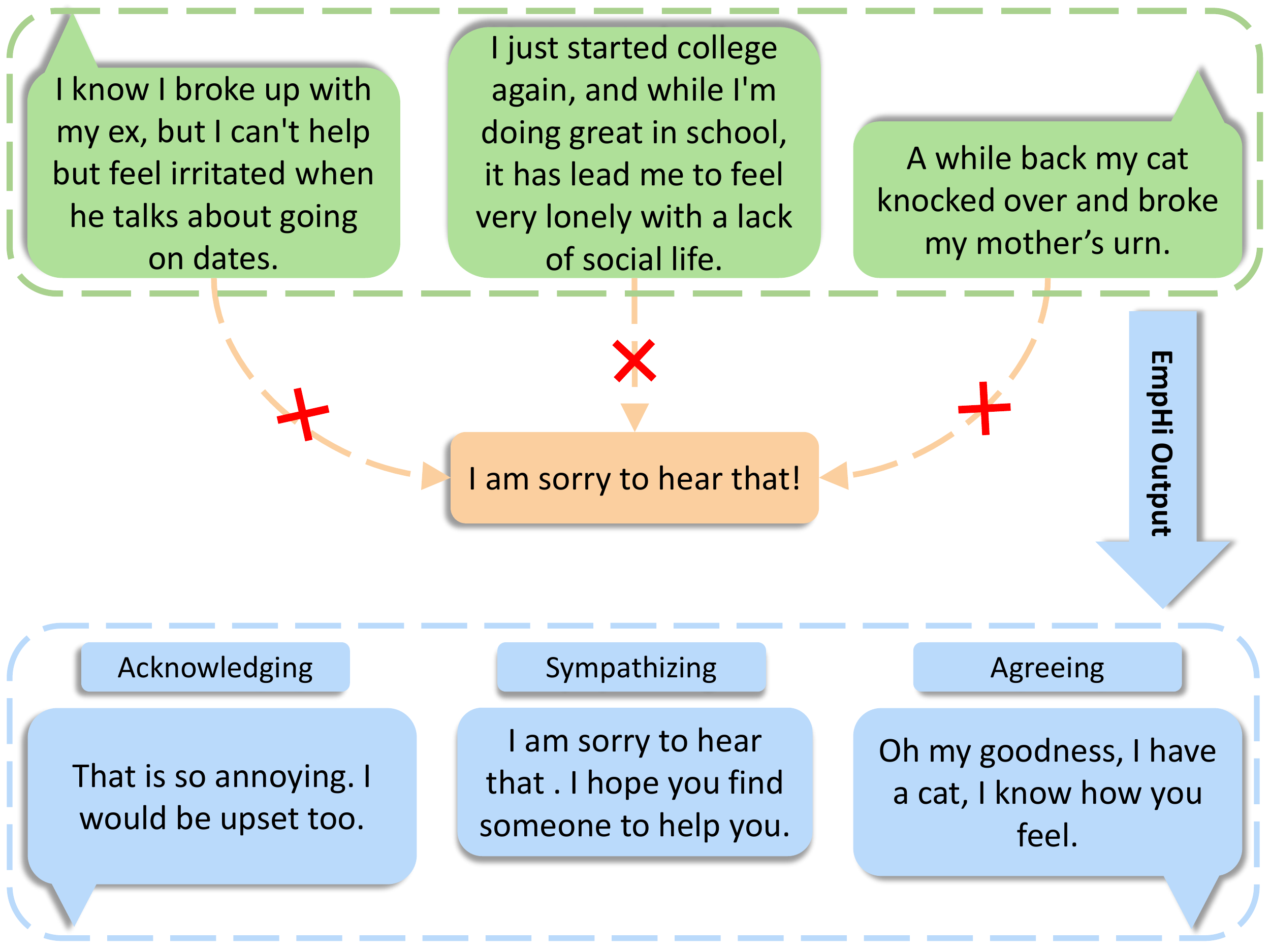}
    \caption{EmpHi generates empathetic responses with human-like empathetic intents (text in blue box), while existing empathetic dialogue models generate responses with contextually irrelevant and monotonous empathy (text in orange box).}
    \label{fig:intro}
\end{figure}

\begin{figure}
    \centering
    \includegraphics[width=\linewidth]{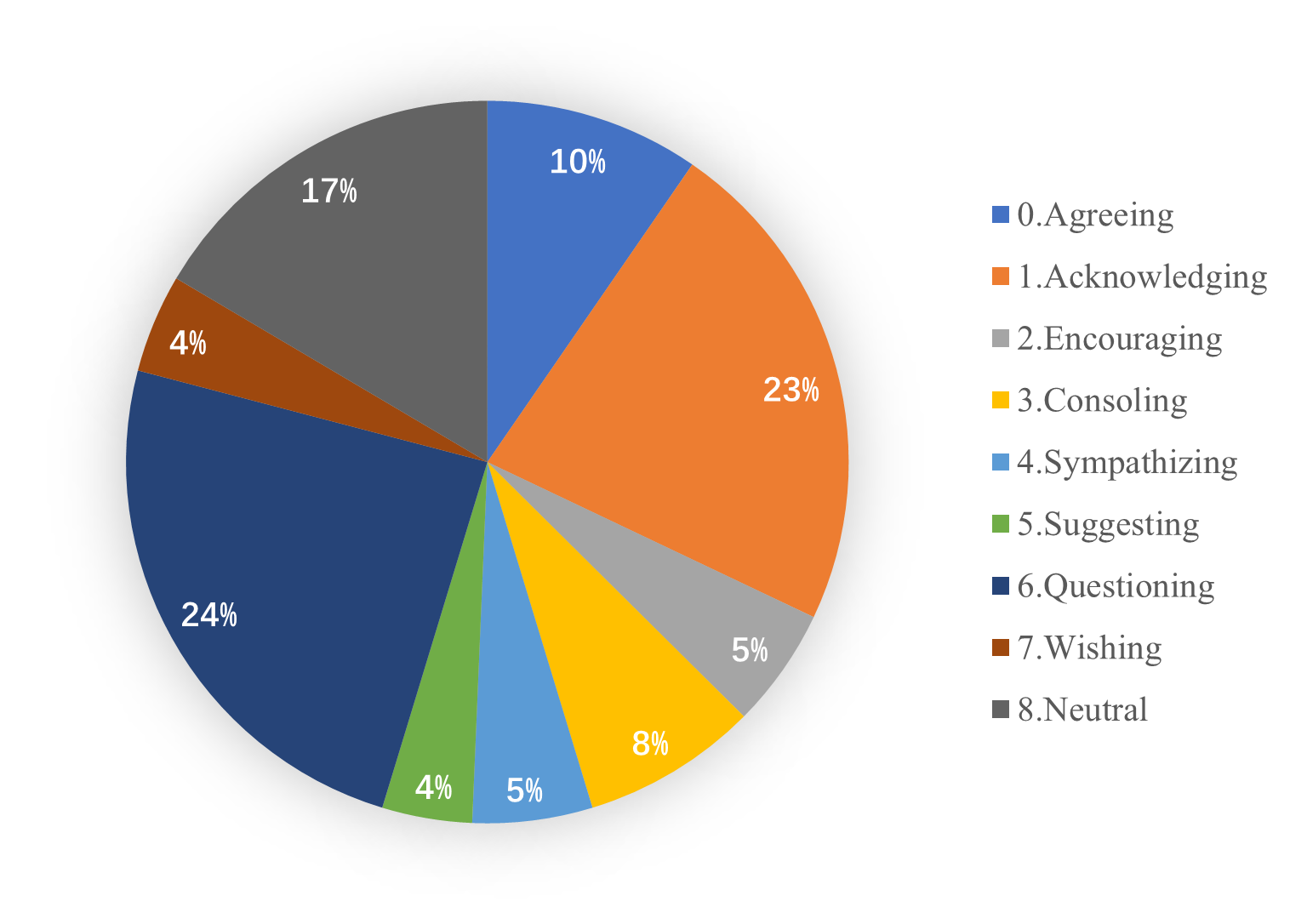}
    \caption{Empathetic intent distribution of human in empathetic conversation.}
    \label{fig:intent_dist}
\end{figure}

Existing empathetic dialogue models pay more attention to the emotion-dependent response generation \cite{lin-etal-2019-moel, majumder-etal-2020-mime}. 
However, using emotion alone to generate responses is coarse-grained, and the model needs to incorporate empathetic intent information.
On the one hand, the speaker often talks with a particular emotion while the listener shows their empathy with specific empathetic intents, e.g., \textit{Acknowledging}, \textit{Agreeing}, \textit{Consoling} and \textit{Questioning} etc \cite{welivita-pu-2020-taxonomy}.
On the other hand, see in Figure \ref{fig:intro}, 
when the user expresses sadness, existing models tend to generate sympathetic responses like "I'm sorry to hear that." However, empathy is not the same as sympathy, so the models should not only generate responses with \textit{Sympathizing} intent. We demonstrate this phenomenon elaborately with a quantitative evaluation in Section \ref{fig:intent_dist}.
In real life situation, humans could reply with various empathetic intents to the same context which depends on personal preference. 
For example, given a context, "I just failed my exam", an individual may respond "Oh no, what happened?" with \textit{Questioning} intent to explore the experience of the user, or "I understand this feeling, know how you feel" with \textit{Agreeing} intent. 
These types of empathy are more relevant, interactive, and diverse.

To address the above issues, we propose a new framework to generate empathetic responses with human-like empathetic intents (EmpHi) which could generate responses with various empathetic intents, see examples in Figure \ref{fig:intro}. 
Specifically, EmpHi learns the empathetic intent distribution with a discrete latent variable and adopts intent representation learning in the training stage. 
During the generation process, EmpHi first predicts a potential empathetic intent and then combines both implicit and explicit intent representation to generate a response corresponding to the predicted intent. 
Our main contributions are:
\begin{itemize}
    
    \item We discover and quantify the severe bias of empathetic intents between existing empathetic dialogue models and humans. 
    This finding will lead subsequent researchers to pay more attention to fine-grained empathetic intents.

    \item To address the above problem, we propose EmpHi, which generates responses with human-like empathetic intents.
    Experiments have proved the effectiveness of our model through the significant improvement in both automatic and human evaluation.
    
    \item According to the quantitative evaluation and analysis, EmpHi successfully captures humans' empathetic intent distribution, and shows high interpretability in generation process.
\end{itemize}
\section{Related Work}

\textbf{Empathetic Response Generation.} Providing dialogue agents the ability to recognize speaker feelings and reply according to the context is challenging and meaningful. \newcite{rashkin-etal-2019-towards} propose the \textbf{EmpatheticDialogues} for empathetic response generation research. 
Most subsequent empathetic conversation researches are evaluated on this dataset, including ours. 
They also propose Multitask-Transformer, which is jointly trained with context emotion classification and response generation. 
To further capture the fine-grained emotion information, \newcite{lin-etal-2019-moel} introduce MoEL, a transformer with a multi-decoder. Each of them is responsible for the response generation of one specific emotion. 
\newcite{majumder-etal-2020-mime} propose MIME, which mimics the speaker emotion to a varying degree.

All these models focus on emotion-dependent empathetic response generation, whereas we pay more attention to the empathetic intents and propose to generate a response that is not only emotionally appropriate but also empathetically human-like.

\textbf{One-to-many Response Generation.} Given dialogue history, there could be various responses depends on personal preference. \newcite{zhao-etal-2017-learning} propose to learn the potential responses with continuous latent variable and maximize the log-likelihood using Stochastic Gradient Variational Bayes (SGVB) \cite{kingma2013auto}. 
To further improve the interpretability of response generation, \newcite{zhao-etal-2018-unsupervised} propose to capture potential sentence-level representations with discrete latent variables. MIME \cite{majumder-etal-2020-mime} introduces stochasticity into the emotion mixture for various empathetic responses generation.

Different from the previous works, we propose a discrete latent variable to control the empathetic intent of response and achieve intent-level diversity.

\begin{figure*}
    \centering
    \includegraphics[width=\textwidth]{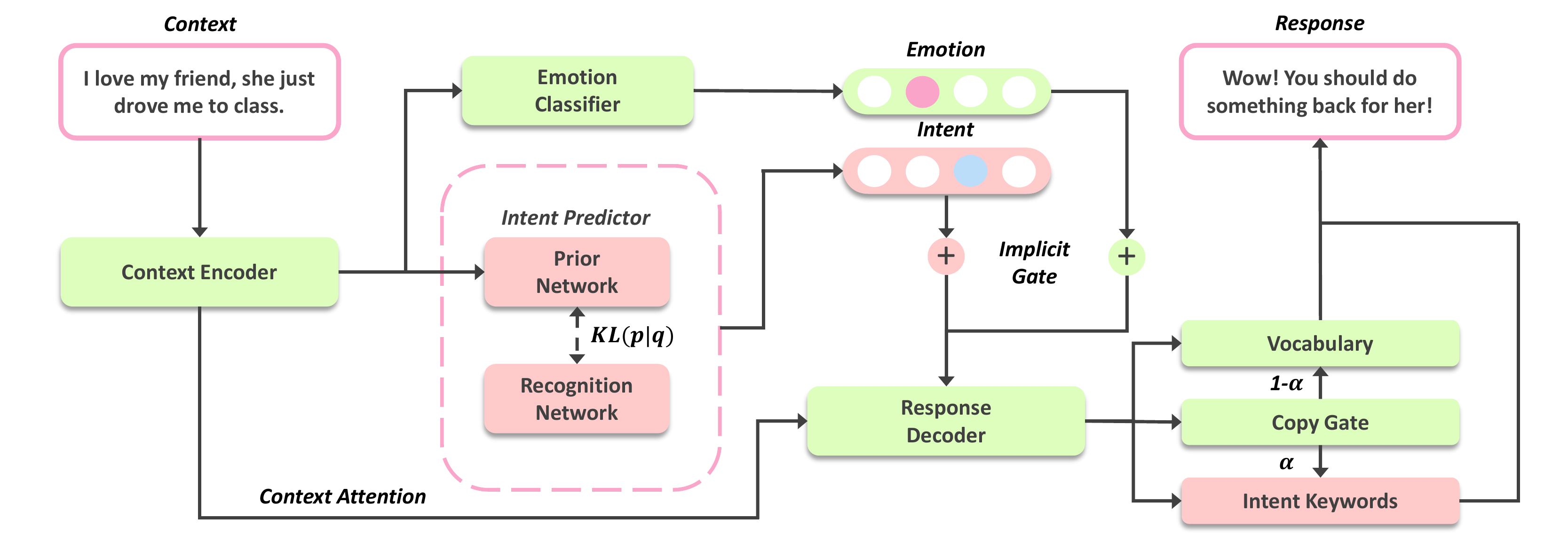}
    \caption{The architecture of EmpHi, which consists of a context encoder, an emotion classifier, a prior network (intent predictor), a recognition network, and a response decoder with copy mechanism.}
    \label{fig:pipeline}
\end{figure*}

\begin{figure}
    \centering
    \includegraphics[width=\linewidth]{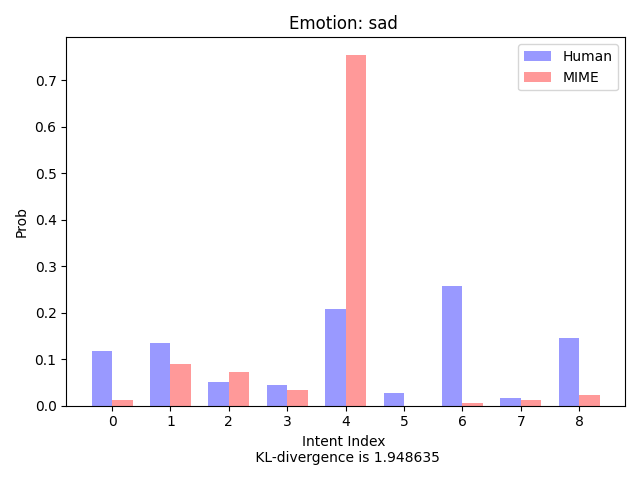}
    \caption{Empathetic intent distribution of human and MIME (sad emotion), the intent index represents the same intent as in Figure \ref{fig:intent_dist}.}
    \label{fig:intent}
\end{figure}

\section{Empathetic Expression Bias}
\label{sec:intent_bias}
Although existing empathetic conversational methods have shown promising progress, we reveal there is a severe bias of empathetic expression between them and humans according to quantitative evaluation.

Empathy plays a vital role in human conversation,
\newcite{welivita-pu-2020-taxonomy} make a taxonomy of empathetic intents and calculate the frequency of each intent in \textbf{EmpatheticDialogues} \cite{rashkin-etal-2019-towards}.
As shown in Figure \ref{fig:intent_dist}, humans show their empathy naturally by \textit{Questioning}, \textit{Acknowledging}, and \textit{Agreeing} intents etc.

However, there are no empirical experiments have shown \textit{how empathetic dialogue models express their empathy?} 
To further study, we finetune a BERT classifier \cite{devlin-etal-2019-bert} on the released \textbf{EmpatheticIntents}$\footnote{\url{https://github.com/anuradha1992/EmpatheticIntents}
}$ dataset \cite{welivita-pu-2020-taxonomy}. 
Our classifier achieves $87.75\%$ accuracy in intent classification and we apply it to identify the empathetic intents of responses generated by the state-of-the-art empathetic dialogue model MIME \cite{majumder-etal-2020-mime}. 
As shown in Figure \ref{fig:intent},
the severe empathetic intent distribution bias emerges while comparing humans to MIME.
Given context with sad emotion, existing models usually generate "I am sorry to hear that" with \textit{Sympathizing} intent, which is not human-like and contextually relevant.
In addition, we can tell that the empathetic expression of MIME is monotonous. 
We also quantify the intent distribution of other empathetic dialogue models in the Appendix \ref{app:gap}. The results are similar to Figure \ref{fig:intent}.

We believe this phenomenon is caused by that existing models only generate responses according to context emotion and lack fine-grained empathetic intent modeling.
Therefore, we propose EmpHi, which generates empathetic responses with human-like empathetic intents.

\section{EmpHi Method}

\subsection{Task Definition and Overview}
Given the context, $C=[c_1, c_2, \cdots, c_m]$, which consists of $m$ words for single or multiple utterances. We aim to generate empathetic response, $X=[x_1, x_2,\cdots, x_n]$, with human-like empathetic intent. The whole model architecture is shown in Figure \ref{fig:pipeline}.





\begin{figure}[t]
    \centering
    \includegraphics[width=\linewidth]{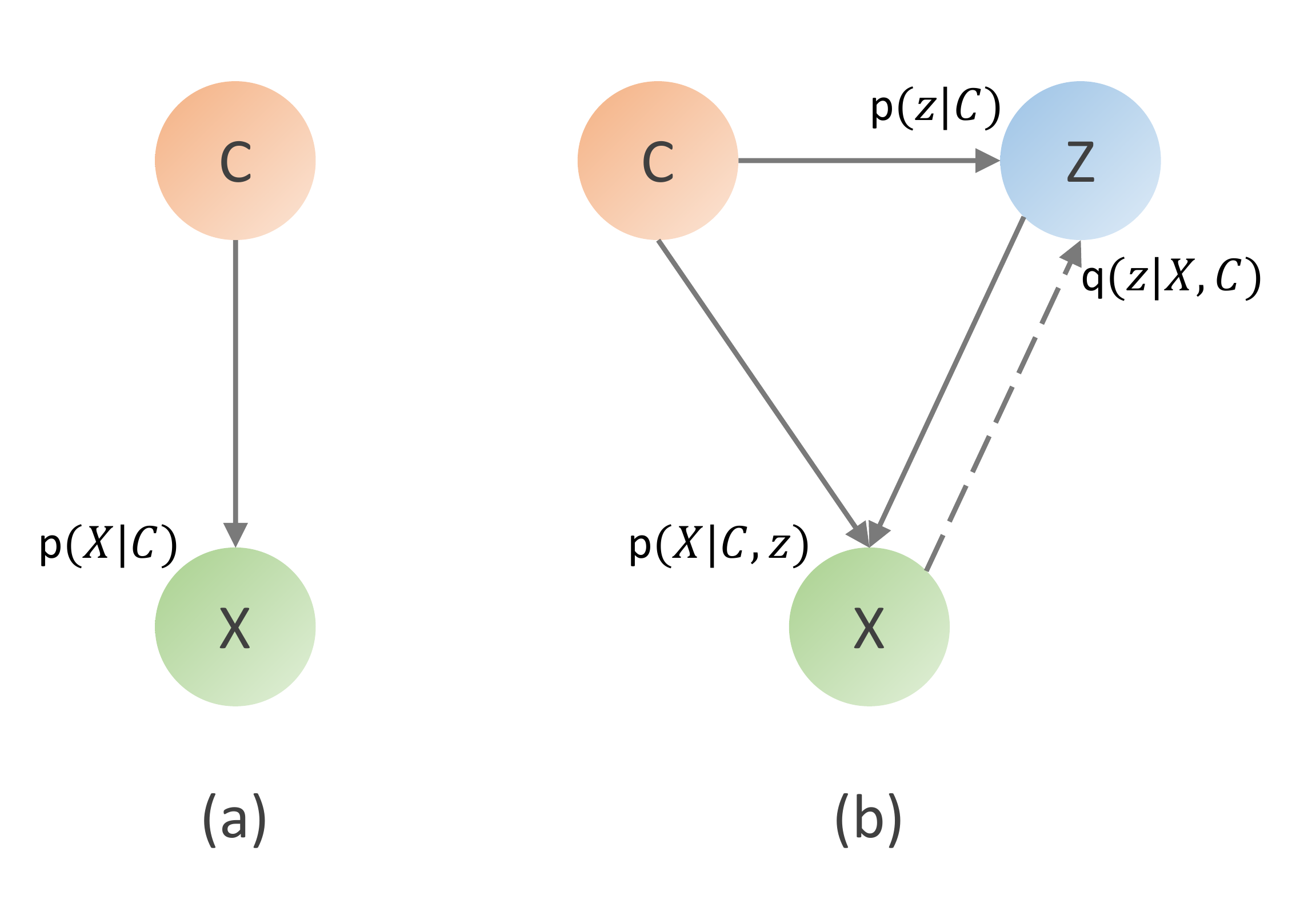}
    \caption{An illustration of the difference between existing empathetic dialogue models (a) and EmpHi (b).}
    \label{fig:graph}
\end{figure}

EmpHi learns the potential empathetic intent distribution with a latent variable $z$, which could be seen in Figure \ref{fig:graph}. 
Conditional Variational AutoEncoder (CVAE) \cite{yan2015attribute2image, zhao-etal-2017-learning, gu2018dialogwae} is trained to maximize the conditional log likelihood, $\log p(X|C)$, which involves an intractable marginalization over $z$.
We train the CVAE efficiently with \textit{Stochastic Gradient Variational Bayes} (SGVB) \cite{kingma2013auto} by maximizing the variational lower bound of the log likelihood: 
\begin{equation}
    \begin{aligned}
        \log p(X|C) \geq &\mathbf{E}_{q(z|X,C)}[\log p(X|C,z)] \\
                &- \mathbf{KL}(q(z|X,C)||p(z|C)),
    \end{aligned}
    \label{eq:cvae}
\end{equation}
$p(X|C,z)$ denotes response reconstruction probability, $q(z|X,C)$ is recognition probability and $p(z|C)$ is prior probability.
Our method mainly consists of three aspects: 
\begin{itemize}
    \item To capture the explicit relationship between the latent variable and the intent, we propose an intent representation learning approach to learn the intent embeddings.
    \item We construct an intent predictor to predict potential response intent using contextual information and then use this intent for guiding the response generation.
    \item During the generation process, EmpHi combines both implicit intent embedding and explicit intent keywords to generate responses corresponding to the given intents. 
\end{itemize}

\subsection{Learning Intent Representation}
\label{sec:intent_rep}
To achieve more interpretability, we choose a discrete latent variable that obeys categorical distribution with nine categories, each corresponding to one empathetic intent.
Directly maximizing Eq.\ref{eq:cvae} would cause two serious problems: the relation between the latent variable and intent is intractable; the \textit{vanishing latent problem} results in insufficient information provided by the latent variable during generation. \cite{bowman-etal-2016-generating, zhao-etal-2017-learning, gu2018dialogwae}. 

To solve the above issues, we separately train a recognition network $q_r(z|X)$ to encourage intent variable $z$ to capture context-independent semantics, which is essential for $z$ to be \textit{interpretable} \cite{zhao-etal-2018-unsupervised}. 
The task of the recognition network is to provide the accurate intent label of the response, which corresponds to an intent embedding.
Then, by maximizing likelihood $p(X|C,z)$, the embedding captures corresponding intent representation automatically.
The recognition network $q_r(z|X)$ does not need additional training. We utilize the BERT intent classifier mentioned above, which achieves $87.75\%$ accuracy in intent classification. In addition, as the sample operation easily brings noise for the intent representation learning when sampling a wrong intent, we use \textup{argmax} operation to avoid the noise, the response reconstruction loss is:
\begin{equation}
   \mathcal{L}_1=-\log p(X|C, z_k), \quad z_k=\mathop{\arg\max}_{z_k} q_r(z_k|X),
\end{equation}
$k \in \{0, 1, 2, \cdots, 8\}$, each integer corresponds to a specific empathetic intent as in Figure \ref{fig:intent_dist}.

\subsection{Intent Predictor and Emotion Classifier}
\label{sec:intent_pre}
The intent predictor is based on the prior network $p_i(z|C)$, which predicts the distribution of response intent by the given context.
During inference, we sample potential intents from this distribution in order to generate human-like empathetic responses.
Specifically, the context is encoded with gated recurrent units (GRU) \cite{chung2014empirical}:
\begin{equation}
    h_t = \mathbf{GRU}(h_{t-1}, E(c_t)),
\end{equation}
where $h_t$ is the hidden state of GRU encoder, $E(c_t)$ denotes the word embedding of the $t$-th word in context, we use $h_m$ as context embedding, then the prior network is:
\begin{equation}
    \begin{aligned}
        p_i(z|C)=\mathbf{Softmax}(\mathbf{FFN_z}(h_m)),
    \end{aligned}
\end{equation}
where $\mathbf{FFN}$ represents \textit{Feed-Forward Network} with two layers. The prior intent distribution is supervised by recognition distribution with KL-divergence in Eq.\ref{eq:cvae}:
\begin{equation}
    \begin{aligned}
        \mathcal{L}_2 &=  \mathbf{KL}(q_r(z|X)||p_i(z|C)) \\
        &= \sum_{k=1}^K q_r(z_k|X)\log \frac{q_r(z_k|X)}{p_i(z_k|C)}.
    \end{aligned}
\end{equation}
Since the context emotion is proved to be beneficial to empathetic dialogue generation \cite{rashkin-etal-2019-towards, lin-etal-2019-moel, majumder-etal-2020-mime}, we also employ an emotion classifier to classify the emotion of context:
\begin{equation}
    \begin{aligned}
        \mathcal{P} &= \mathbf{Softmax}(\mathbf{FFN_e}(h_m))], \\
        p_{e_i} &= \mathcal{P}[i]
    \end{aligned}
\end{equation}
Given the ground truth emotion label $\mathbf{e_t}$, the emotion classifier is trained with cross-entropy loss:
\begin{equation}
    \mathcal{L}_3 =-\log p_{\mathbf{e_t}}.
\end{equation}

\subsection{Response Generator}
\label{sec:intent_gen}
As for the response generation $p(X|C,z)$, we consider implicit intent embedding for the high-level abstraction of an intent.
In addition, we also introduce intent keywords for explicitly utilizing intent knowledge during the generation process.

\textbf{Implicit.}
To generate response with an empathetic intent, the most intuitive approach is taking the intent embedding as additional input to decoder during the generation process.
We also consider emotion embedding as traditional empathetic dialogue models:
\begin{equation}
    s_t = \mathbf{GRU}(s_{t-1}, [E(x_{t-1}); v(z); v(e); c_{att}]),
\end{equation}
where $s_t$ is the state of GRU decoder, $c_{att}$ denotes the context attention value which contains key information of context \cite{bahdanau2014neural}. $v(z)$ is intent embedding and $v(e)$ is emotion embedding, both will not change during the generation process. However, this may sacrifice grammatical correctness \cite{zhou2018emotional, ghosh-etal-2017-affect}. 
Therefore we add a gate operation to capture intent and emotion dynamically:
\begin{equation}
    \begin{aligned}
        \textup{Input} &= \mathbf{FFN_i}([E(x_t); c_{att}; s_t]), \\
        \textup{Gate} &= \mathbf{Sigmoid}(\textup{Input}), \\
        \Bar{v}(z) &= \textup{Gate} \odot v(z),
    \end{aligned}
\end{equation}
where $\odot$ represents element-wise product. Each time step, the intent representation is used appropriately according to current word, state, and context value. The gate operation for emotion is the same as above.

\textbf{Explicit.} The empathetic expression is quite distinct over vocabularies, e.g., `know', `understand', `agree', are indicative of the empathetic intent \textit{Agreeing}.
Therefore, we employ the copy mechanism to explicitly utilize intent keywords for intent conditional generation. See in Appendix \ref{app:intent} for more details about intent keywords .
\begin{equation}
    \begin{aligned}
        \alpha_t &= \mathbf{Sigmoid}(v_s^\top s_t), \\
        p(x_t=w_g) &= \mathbf{Softmax}(W_g s_t), \\
        p(x_t=w_i) &= \mathbf{Softmax}(W_i s_t), \\
        p(x_t) &= (1-\alpha_t) \cdot p(w_g) + \alpha_t \cdot p(w_i),
    \end{aligned}
\end{equation}
where $\{s_t, v_s\} \in \mathcal{R}^{d \times 1}$, $\{W_g, W_i\} \in \mathcal{R}^{V \times d}$, $d$ is hidden size and $V$ denotes the vocabulary size. 
The copy rate $\alpha_t$ is used to balance the choice between intent keywords and generic words, it is trained with binary cross entropy loss:
\begin{equation}
    \mathcal{L}_4 = \sum_{t=1}^{n} q_t \cdot \log \alpha_t + (1-q_t) \cdot \log (1-\alpha_t),
\end{equation}
$n$ is the amount of words in response, $q_t \in \{0, 1\}$ indicates that whether $x_t$ is a intent keyword.

\subsection{Loss Function}
To summarize, the total loss is:
\begin{equation}
    \mathcal{L} = \lambda_1\mathcal{L}_1 + \lambda_2\mathcal{L}_2 + \lambda_3\mathcal{L}_3 + \lambda_4\mathcal{L}_4,
\end{equation}
In order to join all losses with weighting method, we add 4 hyperparameters in total loss, $lambda_i$, where each $lambda_i$ is corresponding to $L_i$.
$\mathcal{L}_1, \mathcal{L}_2, \mathcal{L}_3, \mathcal{L}_4$ denote the losses of response reconstruction, intent prediction, emotion classification and copy rate prediction respectively.

\section{Experiments}
\subsection{Dataset}
We evaluate our method and compare with others on \textbf{EmpatheticDialogues}$\footnote{\url{https://github.com/facebookresearch/EmpatheticDialogues}
}$ \cite{rashkin-etal-2019-towards} which contains $25k$ open domain dialogues. Follow the same setting as the authors of this dataset, the proportion of train/validation/test data is $8:1:1$. Each dialogue consists of at least two utterances between a speaker and listener. There are $32$ emotion situations in total, which are uniformly distributed.

\subsection{Baselines}
We compare our model with the three latest empathetic conversational models:
\begin{itemize}
    \item \textbf{Multitask Transformer (Multi-TRS).} A transformer model trained by the response generation task and the context emotion classification task \cite{rashkin-etal-2019-towards}.
    \item \textbf{Mixture of Empathetic Listeners (MoEL).} An enhanced transformer model with $32$ emotion-specific decoders to respond appropriately for each emotion \cite{lin-etal-2019-moel}. 
    \item \textbf{MIMicking Emotions for Empathetic Response Generation (MIME).} The state-of-the-art empathetic dialogue model allows the generator to mimic the context emotion to a varying degree based on its positivity, negativity, and content. Furthermore, they introduce stochasticity into the emotion mixture and achieves one-to-many generation \cite{majumder-etal-2020-mime}.
\end{itemize}

\subsection{Evaluation}
\subsubsection{Automatic Metrics}
\begin{itemize}
    \item \textbf{BLEU.} We choose BLEU \cite{papineni-etal-2002-bleu} for relevance evaluation which measures the $n$-gram overlaps with reference and compute BLEU scores for $n\leq4$ using smoothing techniques \cite{chen-cherry-2014-systematic}. Since the state-of-art model MIME and ours are both one-to-many generators, we calculate BLEU recall and BLEU precision \cite{zhao-etal-2017-learning, gu2018dialogwae}. For each test case, we sample $5$ responses from latent space and use greedy search for MIME and EmpHi, use beam search for MoEL and Multitask-Transformer.
    
    \item \textbf{Distinct.} Distinct \cite{li-etal-2016-diversity} is a widely used metric for diversity evaluation. Specifically, we compute the number of distinct unigrams (Distinct-1) and bigrams (Distinct-2), then scale them by the total number of unigrams and bigrams.  
\end{itemize}

\subsubsection{Human Ratings}
First, we randomly sample $100$ dialogues and their corresponding generations from the three baseline models and EmpHi. Then, we invite five volunteers with master degrees to do the human evaluation. The annotators mark each response from 1 to 5 for empathy, relevance, and fluency.

To clarify the marking criteria, we provide an explanation for each metric:
\begin{itemize}
\item \textbf{Empathy.} Whether the response shows that the listener understands and shares the speaker's feeling. Can the listener imagine what it would be like in the speaker's situation?

\item \textbf{Relevance.} Whether the response is relevant to the context.

\item \textbf{Fluency.} Whether the response is easy to read and grammatically correct.
\end{itemize}

\subsubsection{Human A/B Test}
Following \cite{lin-etal-2019-moel, majumder-etal-2020-mime}, we construct this evaluation task to directly compare our model with each baseline. We randomly sample $100$ dialogue responses from \textit{EmpHi} vs \{\textit{Multitask-Trans}, \textit{MoEL}, \textit{MIME}\}. Given randomly ordered responses from above models, four annotators select the better response, or $tie$ if they think the two responses have the same quality. The average score of four results is calculated, and shown in Table \ref{fig:intent_ours}.

\begin{table*}[ht]
\small
\centering
\begin{tabular}{c|c|c|c|c|ccc|cc}
\toprule
\multirow{2}{*}{Methods} & \multirow{2}{*}{\#Params.} & \multirow{2}{*}{Empathy} & \multirow{2}{*}{Relevance} & \multirow{2}{*}{Fluency} & \multicolumn{3}{c|}{BLEU} & \multicolumn{2}{c}{Distinct}                      \\ \cline{6-10}
\multicolumn{1}{c|}{}    & \multicolumn{1}{c|}{}      & \multicolumn{1}{c|}{}    & \multicolumn{1}{c|}{}      & \multicolumn{1}{c|}{}    & \multicolumn{1}{c}{P} & \multicolumn{1}{c}{R} & \multicolumn{1}{c|}{F1} &     \multicolumn{1}{c}{D-1} & \multicolumn{1}{c}{D-2} \\ \hline \hline
Multitask-Trans & 20M  & 2.68            & 2.58             & 3.60            & 0.3072          & 0.4137          & 0.3526          & 0.4123 & 1.1390 \\ \hline
MoEL            & 21M  & 3.18            & 3.18             & 3.95            & 0.3032          & 0.3614          & 0.3298          & 0.8473 & 4.4698 \\ \hline
MIME            & 18M  & 2.89            & 2.90             & 3.77            & 0.3202          & 0.3278          & 0.3240          & 0.3952 & 1.3299 \\ \hline
EmpHi            & \textbf{11M}  & \textbf{3.48}   & \textbf{3.66}    & \textbf{4.34}   & \textbf{0.3207} & \textbf{0.4723} & \textbf{0.3820} & \textbf{1.1188} & \textbf{5.3332} \\ \hline
Human           & -    & 4.04            & 4.40             & 4.56            & -               & -               & -               & 7.0356 & 43.2174 \\
\bottomrule
                      
\end{tabular}
\caption{Automatic evaluation between EmpHi and other models. All results are the mean of 5 runs for fair comparison. D-1.\&2. are magnified 100 times for each model.}
\label{tab:main}
\end{table*}

\subsection{Implement Detail}
For MIME$\footnote{\url{https://github.com/declare-lab/MIME}
}$ \cite{majumder-etal-2020-mime} and MoEL$\footnote{\url{https://github.com/HLTCHKUST/MoEL}
}$ \cite{lin-etal-2019-moel}, we reproduce their results using their open-source codes and their default hyperparameters. 
According to the log-likelihood in the validation dataset for Multitask-Transformer, we use grid search for the best head number, layer number, and feed-forward neural network size. The best set is 2, 10, and 256, respectively. 
EmpHi uses a two-layer Bi-GRU as the encoder and a two-layer GRU as the decoder, $\lambda$ is set as $[1, 0.5, 0.5, 1]$ respectively. 
All the feed-forward neural networks in EmpHi have two layers, $300$ hidden units and ReLU activations. 
For the sake of fairness, we use pretrained Glove vectors \cite{pennington-etal-2014-glove} with 300 dimensions as the word embedding for all models, the batch size is $16$, and the learning rate is set to $1e^{-4}$.

\begin{table}[]
\small
\centering
\begin{tabular}{c|c|c|c}
\toprule
Methods                  & Win      & Loss      & Tie     \\ \hline \hline
EmpHi vs Multitask-trans & \textbf{56.5}\%   & 21.5\%    & 22.0\%  \\ \hline  
EmpHi vs MoEL            & \textbf{45.0}\%   & 28.5\%    & 26.5\%  \\ \hline
EmpHi vs MIME            & \textbf{53.0}\%   & 27.0\%    & 20.0\%  \\ \hline
\bottomrule
                      
\end{tabular}
\caption{Results of Human A/B test.}
\label{tab:ab}
\end{table}

\section{Results and Discussions}

\subsection{Results Analysis}

In this section, we mainly testify:
\begin{itemize}
    \item human-like empathetic intent boost EmpHi's performance in terms of empathy, relevance, and diversity.
    \item EmpHi successfully captures the empathetic intent distribution of humans.
\end{itemize}
\subsubsection{Human Evaluation}

As shown in Table \ref{tab:main}, EmpHi outperforms all baselines in terms of empathy, relevance, and fluency. The most distinct improvement is 15.1\% on relevance because our model does not only depends on the speakers' emotion, but also makes use of the empathetic intents, which are contextually relevant. It is worth noting that empathy is the primary metric in empathetic dialogue generation. 
EmpHi outperforms the previous SOTA on empathy by 9.43\%, which directly indicates that human-like empathetic intents are beneficial to the empathy ability of the dialogue model.

Last but not least, a decent fluency score proves that our generated response could be understood by humans easily, where our model has an improvement of 9.87\% from MoEL. In addition, the human A/B test results in Table \ref{tab:ab} also confirm that the responses from our model are preferable to baselines. Overall, 
EmpHi successfully generates empathetic, relevant, and fluent responses.



\begin{figure}
    \centering
    \includegraphics[width=\linewidth]{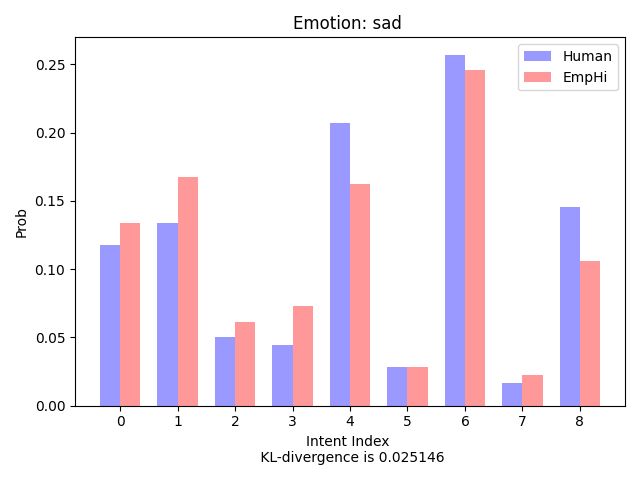}
    \caption{Empathetic intent distribution of human and EmpHi (sad emotion), the intent index represents the same intent as in Figure \ref{fig:intent_dist}.}
    \label{fig:intent_ours}
\end{figure}

\begin{table}[t]
\small
\centering
\begin{tabular}{c|ccc|c}
\toprule
\multirow{2}{*}{Methods} & \multicolumn{3}{c|}{BLEU} & \multirow{2}{*}{ACC} \\ \cline{2-4}
\multicolumn{1}{c|}{}    & \multicolumn{1}{c}{P} & \multicolumn{1}{c}{R} & \multicolumn{1}{c|}{F1}  & \multicolumn{1}{c}{} \\ \hline \hline
EmpHi                     & 0.3207 & \textbf{0.4723} & \textbf{0.3820} & \textbf{26.8}\% \\ \hline
EmpHi w/o intent & 0.3105 & 0.4049 & 0.3515 & 21.9\%\\ \hline
EmpHi w/o gate & 0.3138 & 0.4634 & 0.3742 & 25.3\%\\ \hline
EmpHi w/o copy & \textbf{0.3215} & 0.4704 & 0.3820 & 25.9\%\\
\bottomrule
                      
\end{tabular}
\caption{Results of ablation study.}
\label{tab:ablation}
\end{table}

\subsubsection{Automatic Evaluation}
As seen in Table \ref{tab:main}, the automatic evaluation is consistent with human evaluation. 
The BLEU recall and F1 score are improved by $14.2\%$ and $8.34\%$, respectively. 
However, we only have a slight improvement on BLEU precision, which is similar to \cite{zhao-etal-2017-learning, gu2018dialogwae} because the precision is penalized when the model generates diverse responses. 
Also, the distinct value of unigrams and bigrams are $32.04\%$ and $19.32\%$ higher than the previous SOTA, respectively. 
As shown in Figure \ref{fig:intent} and Figure \ref{fig:intent_ours}, the empathy intents of EmpHi's responses are more diverse than existing models, so the distinct scores improve significantly.
Our method enhances the relevance and diversity simultaneously, which proves the effectiveness of human-like intent in empathetic response generation. 


\begin{table*}[]
    \centering
    \scalebox{0.8}{
    \begin{tabular}{l|p{14cm}}
    \toprule
    \textbf{Emotion}         & Nostalgic \\ \hline
    \textbf{Context}         & When my wife and i started dating, our second date ended up lasting like 12 hours.  \\ \hline
    \textbf{Reference}       & That is a pretty long date ! Where did you guys go? \\ \hline
    \textbf{Multitask-Trans} & That is so sad, I'm sorry to hear that. \\ \hline
    \textbf{MoEL}            & That is great! I am sure she will be fine. \\ \hline
    \textbf{MIME}            & That is a good thing to do. I am sure you will do great.  \\ \hline
    \textbf{EmpHi}            & That 's awesome ! How long have you been together ?\\ \hline \hline
    
    \textbf{Emotion}         & Angry \\ \hline
    \textbf{Turn-1}          & I just moved to this neighborhood and some dumb criminals shot one of my neighbors and ran into the woods!  \\ \hline
    \textbf{Turn-2}          & That's not good. Do you own a gun? \\ \hline
    \textbf{Turn-3}          & I do! I want to be able to protect my son. \\ \hline
    \textbf{Reference}       & That is always number one goal. \\ \hline
    \textbf{Multitask-Trans} & What did you do ? \\ \hline
    \textbf{MoEL}            & That is a good idea . \\ \hline
    \textbf{MIME}            & I am sorry to hear that. I hope you get it !    \\ \hline
    \textbf{EmpHi}           & Maybe you should go to the police.\\ 
    \bottomrule
    
    \end{tabular}}
    
    \caption{Comparison among EmpHi and other empathetic dialogue models.}
    \label{tab:cas2}
\end{table*} 

\subsubsection{Empathetic Intent Distribution}
We apply the same approach in Section \ref{sec:intent_bias} and quantify the empathetic intent distribution of EmpHi’s responses to prove that EmpHi accurately captures humans' empathetic intent distribution.
Comparing Figure \ref{fig:intent} and Figure \ref{fig:intent_ours}, the difference between them illustrates that our model successfully reduces the bias of empathetic expression. The KL-divergence of intent distributions between models and humans are  $\textbf{0.025}$ for EmpHi, $1.949$ for MIME, $1.545$ for MoEL, and $4.570$ for Multitask-Transformer (See in Appendix \ref{app:gap}).

\subsection{Ablation Study}
We evaluate each component of EmpHi using BLEU and ACC, where ACC indicates the accuracy of predicted empathethetic intent of generated response. Since each conversation could have multiple empathetic responses, the ACC of 26.8\% is pretty ideal.
As seen in Table \ref{tab:ablation}, there is a dramatic drop in the performance of EmpHi without any intent information (both implicit embedding and explicit keywords). Therefore, this proves the effectiveness of empathetic intents and the intent representation learning approach.
As for implicit gate control, it improves both response quality and intent accuracy since it helps EmpHi dynamically capture intent information during generation. Same conclusion has been made in \cite{zhou2018emotional}.
The copy mechanism provides EmpHi the ability to explicitly use intent keywords and thus contributes to the intent accuracy.


\begin{figure}[]
    
    \centering
    \includegraphics[width={\linewidth}]{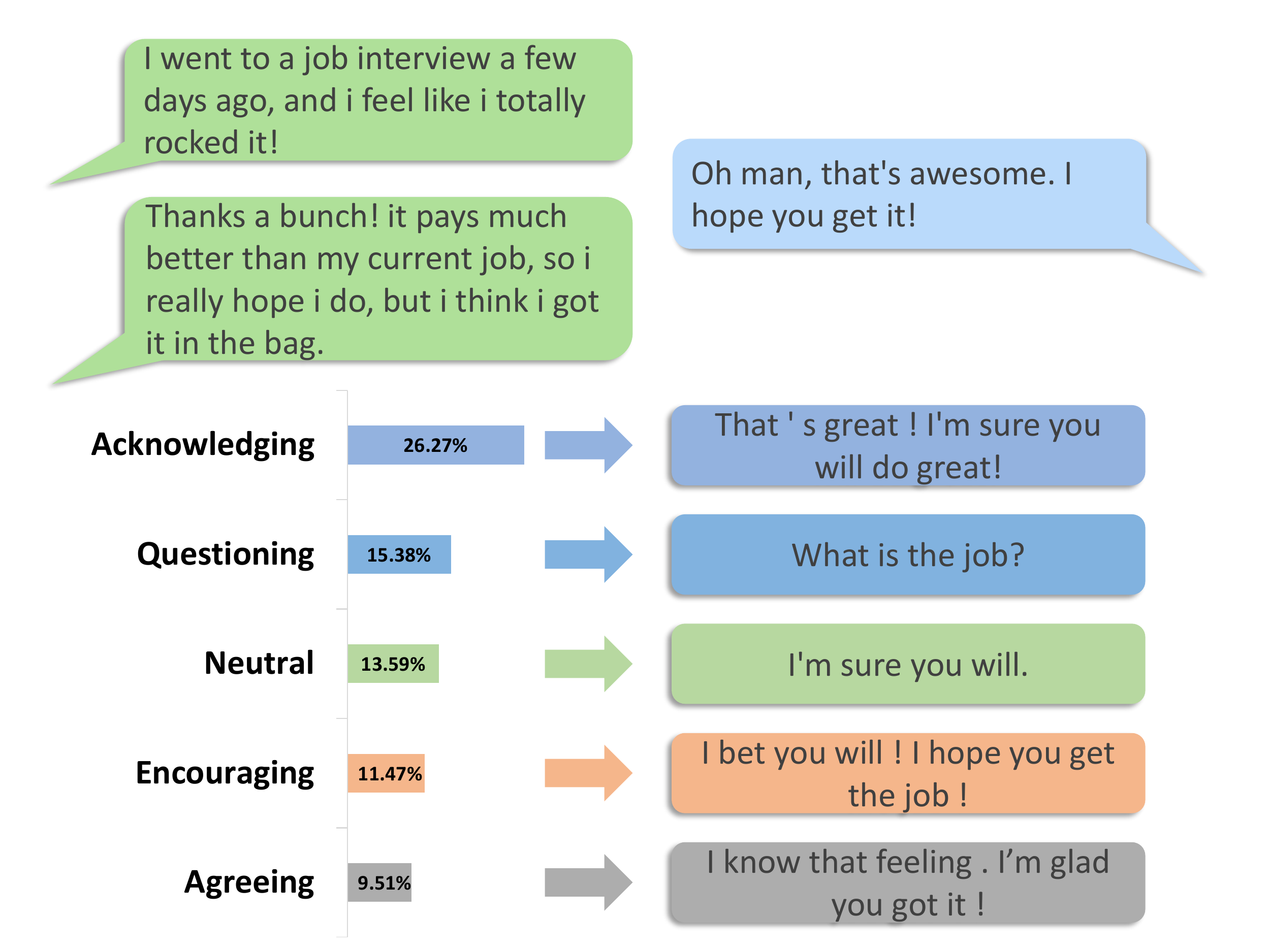}
    \caption{Case study of EmpHi.}
    \label{fig:cas1}
\end{figure}

\subsection{Case Study}
\textbf{Intent-level diverse generation.} Through sampling intents in the discrete latent space, EmpHi generates different responses with empathetic intents.
As in Figure \ref{fig:cas1}, the speaker shows an exciting emotion for getting a better job. 
EmpHi generates empathetic yet contextually relevant responses as humans. 
Besides, EmpHi predicts the potential intent distribution and shows successful conditional generation based on the corresponding intents, which improves the interpretability and controllability of empathetic response generation. 
See Appendix \ref{app:err} for error analysis. 

\textbf{Compare with existing models.}
For the first instance in Table \ref{tab:cas2}, even though baseline models show naive empathy in their response, it is hard for the speaker to feel empathy because the response is not relevant to the topic. 
In contrast, EmpHi shows its understanding of the speaker's feelings and asks a relevant question to explore the speaker's experience. 
For second case, all baselines express contextually irrelevant empathy, while EmpHi truly understands the dialogue history and put itself into speaker's situation, then further reply: "Maybe you should go to the police" with the \textit{Suggesting} intent.


\section{Conclusion}

Overall, we reveal the severe bias of empathetic expression between existing dialogue models and humans. To address this issue, this paper proposes EmpHi to generate empathetic responses with human-like empathetic intents. As a result, both automatic and human evaluation prove that EmpHi has a huge improvement on empathetic conversation. According to the anlaysis and case studies, EmpHi successfully learns the emapthetic intent distribution of human and shows high interpretability and controllability during the generation process. We will try large pretrained language models with empathetic intent in our future work.

\section*{Ethical Statement}
Since this paper involves subjects related to human conversation, we have ensured that all the experiments will cause no harm to humans. The dataset EmpatheticDialogues is collected by \cite{rashkin-etal-2019-towards}, all the participants join the data collection voluntarily. Also, the dataset provider filters all personal information and obscene languages. Therefore, we believe that the dataset EmpatheticDialogues used in our experiments are harmless to users, and the model trained on this dataset is not dangerous to humans.

\section*{Acknowledgements}
We thank the anonymous reviewers for their insightful comments and suggestions. This research was supported in part by the National Key Research and Development Program of China (No. 2020YFB1708200), the Guangdong Basic and Applied Basic Research Foundation (No. 2019A1515011387) and the Shenzhen Key Laboratory of Marine IntelliSense and Computation under Contract ZDSYS20200811142605016.

\bibliography{anthology,custom}
\bibliographystyle{acl_natbib}

\clearpage
\appendix

\begin{table*}[]
    \centering
    \scalebox{0.9}{
    \begin{tabular}{l|p{14cm}}
    \toprule
    Intent         & Keywords \\ \hline \hline
    Agreeing         & know, understand, agree, definitely, feel, feeling, exactly, mean, oh, right  \\ \hline
    Acknowledging       & sounds, nice, awesome, like, great, cool, would, oh, must, really \\ \hline
    Encouraging     & hope, well, hopefully, get, good, time, bet, great, goes, soon \\ \hline
    Consoling            & hope, hopefully, get, better, well, soon, time, find, good, things \\ \hline
    Sympathizing            & sorry, hear, oh, am, happened, loss, feel, hope, really, aw  \\ \hline
    Suggesting            & maybe, get, time, could, think, well, next, something, try, go \\ \hline 
    Questioning        & oh, get, go, going, long, kind, like, work, good, maybe \\ \hline
    Wishing        & congratulations, luck, good, wish, best, well, happy, oh, birthday, wow \\ \hline
    Neutral        & good, like, sure, well, time, one, have, people, never, get \\ \bottomrule
    \end{tabular}}
    
    \caption{Keywords for each intent.}
    \label{tab:keyword}
\end{table*} 

\begin{table*}[h]
    \centering
    \scalebox{0.9}{
    \begin{tabular}{l|l|p{13cm}}
    \toprule
    \multicolumn{2}{l|}{Emotion} & Proud \\ \hline
    \multicolumn{2}{l|}{Context}         & My wife was able to get a job with microsoft without going to a four-year college!  \\ \hline
    \multicolumn{2}{l|}{Reference} & That' s pretty amazing, she must be skillful regardless. \\ \hline \hline
    Intent          &  Prob             & Generated Response  \\ \hline
    Agreeing        & 2.91\%             & Oh, wow ! That's amazing ! I know how you feel. \\ \hline
    Acknowledging   & \textbf{40.9\%}   & That's awesome ! I'm glad you're going to be proud of her. \\ \hline
    Encouraging     & 9.98\%            & That's awesome ! I hope you guys have a great job! \\ \hline
    Consoling       & 3.66\%             & Wow, That' s good, i hope you are able to get it .   \\ \hline
    Sympathizing    & 1.43\%             & Oh wow ! I am sorry to hear that. \\ \hline
    Suggesting      & 0.62\%             & That's good ! Maybe she will be fine. \\ \hline
    Questioning     & 28.6\%            & That's great ! How long have you been working ? \\ \hline
    Wishing         & 9.19\%             & Congratulations ! That' s awesome ! congratulations ! \\ \hline
    Neutral         & 2.75\%             & That is a good thing to do, I wish i was a lot of work .  \\
    \bottomrule
    
    \end{tabular}}
    \caption{Error analysis of EmpHi, although most responses are reasonable, there are still some imperfection.}
    \label{tab:error}
\end{table*}

\section{Empathetic Expression Gap}
\label{app:gap}
\begin{figure}[h]
    \centering
    \includegraphics[width=\linewidth]{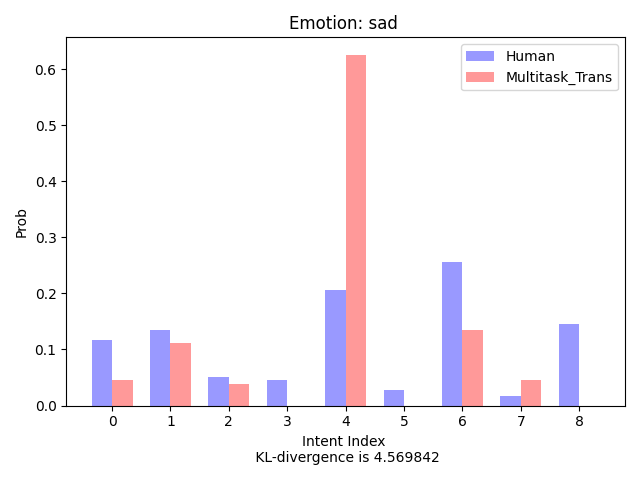}
    \caption{Empathetic intent distribution of human and Multitask-Transformer \cite{rashkin-etal-2019-towards}}
    \label{fig:intent_multitask_trans}
\end{figure}

\begin{figure}[h]
    \centering
    \includegraphics[width=\linewidth]{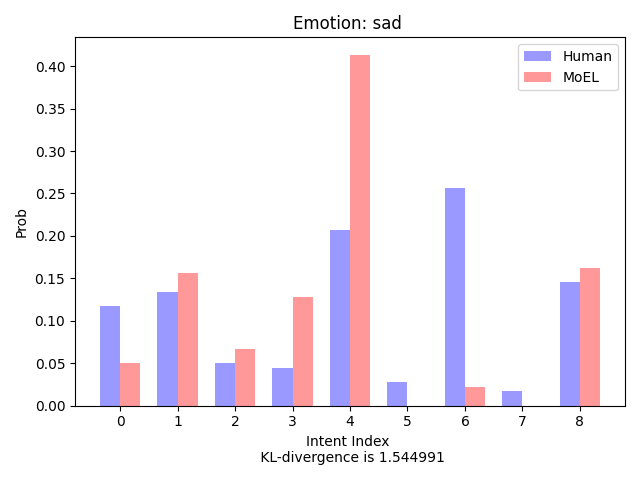}
    \caption{Empathetic intent distribution of human and MoEL \cite{lin-etal-2019-moel}}
    \label{fig:intent_moel}
\end{figure}

For more comprehensive recognization of the severe emathy expression bias between existing empathetic dialogue models and humans, we further quantify the bias of Multitask-Transformer \cite{rashkin-etal-2019-towards} in Figure \ref{fig:intent_multitask_trans} and MoEL \cite{lin-etal-2019-moel} in Figure \ref{fig:intent_moel}, the intent index is consistent with Figure \ref{fig:intent_dist}.
The results are similar with MIME \cite{majumder-etal-2020-mime}, we can see the large intent distribution bias and the monotony of empathetic expression of existing models.

\section{Intent Keywords Collection}
\label{app:intent}
The keywords are retrieved from the training set of \textbf{Empathetic Intents} dataset \cite{welivita-pu-2020-taxonomy} by using TF-IDF method. \textbf{Empathetic Intents} has a training set of 5490 responses, where each intent group has $610$ responses. Based on the labeled intent for each response in the training set, we concatenate all the responses which are in the same group and remove all the stop words. Finally, we apply TF-IDF to obtain the top $k$ keywords for each intent group, we set $k$ to $30$ in our experiments. See Table \ref{tab:keyword} for top ten keywords for each intent.

\section{Error Analysis}
\label{app:err}
Although EmpHi achieves huge improvement in terms of empathy, relevance, and diversity in empathetic dialogue generation, there is still some flaws. At first, the generation task of EmpHi is far difficult than existing models, because it needs to generate response condition on both context and the predicted intent, while other models generate response only condition on the context, therefor the exposure bias of EmpHi is more severe. See in Table \ref{tab:error}, although the predicted intent of EmpHi is the same as reference and its corresponding response is great, EmpHi also gives high probability for \textit{Questioning} intent and the corresponding response is not very contextually relevant, EmpHi knows it is suitable for asking more details to show its caring, but it does not know how to ask under this context, thus EmpHi needs better understanding for context information. We believe this issue could be mitigated when using more dialogue data for pretraining.

\end{document}